\documentclass[runningheads]{llncs}
\usepackage[T1]{fontenc}
\usepackage{graphicx}
\usepackage{booktabs}
\usepackage{amsmath}
\usepackage{amssymb}
\usepackage{amssymb}
\usepackage{orcidlink}
\usepackage{breakcites}

\usepackage{xcolor}
\newif\ifcolor
\colortrue  
\newcommand{\cj}[1]{#1}
\newcommand{\sm}[1]{#1}
\newcommand{\jf}[1]{#1}
\newcommand{\nb}[1]{#1}

\newcommand{\cjj}[1]{\ifcolor\textcolor{black}{#1}\else#1\fi}
\newcommand{\smm}[1]{\ifcolor\textcolor{black}{#1}\else#1\fi}
\newcommand{\jff}[1]{\ifcolor\textcolor{black}{#1}\else#1\fi}

\newcommand{\jung}[1]{\ifcolor\textcolor{black}{#1}\else#1\fi}

\begin{document}
%
\title{\nb{Computer Vision for Objects used in Group Work: Challenges and Opportunities}}
%
%
\author{Changsoo Jung\orcidlink{0000-0002-2232-4300} \and
Sheikh Mannan\orcidlink{0000-0001-9715-4847} \and
Jack Fitzgerald\orcidlink{0009-0008-5604-7920} \and
Nathaniel Blanchard\orcidlink{0000-0002-2653-0873}
}
\authorrunning{C. Jung et al.}

%
\institute{Colorado State University, Fort Collins, CO 80523, USA \\
\email{\{Changsoo.Jung, sheikh.mannan, Jack.Fitzgerald, Nathaniel.Blanchard\}@colostate.edu}
}

%
\maketitle              
\begin{abstract}
 Interactive and spatially aware technologies are transforming educational frameworks, particularly in K-12 settings where hands-on exploration fosters deeper conceptual understanding. However, during collaborative tasks, existing systems often lack the ability to accurately capture real-world interactions between students and physical objects. This issue could be addressed with automatic 6D pose estimation, i.e., estimation of an object's position and orientation in 3D space from RGB images or videos. For collaborative groups that interact with physical objects, 6D pose estimates allow AI systems to relate objects and entities. As part of this work, we introduce FiboSB, a novel and challenging 6D pose video dataset featuring groups of three participants solving an interactive task featuring small hand-held cubes and a weight scale. This setup poses unique challenges for 6D pose because groups are holistically recorded from a distance in order to capture all participants --- this, coupled with the small size of the cubes, makes 6D pose estimation inherently non-trivial. We evaluated four state-of-the-art 6D pose estimation methods on FiboSB, exposing the limitations of current algorithms on collaborative group work. An error analysis of these methods reveals that the 6D pose methods' object detection modules fail. \cjj{We address this by fine-tuning YOLO11-x for FiboSB, achieving an overall $mAP_{50}$ of 0.898.} The dataset, benchmark results, and analysis of YOLO11-x errors presented here lay the groundwork for leveraging the estimation of 6D poses in difficult collaborative contexts.

\end{abstract}

\keywords{6D pose \and collaborative group work \and computer vision.}

\section{Introduction}


 \begin{figure}[t]
  \centering
  \includegraphics[width=\textwidth]{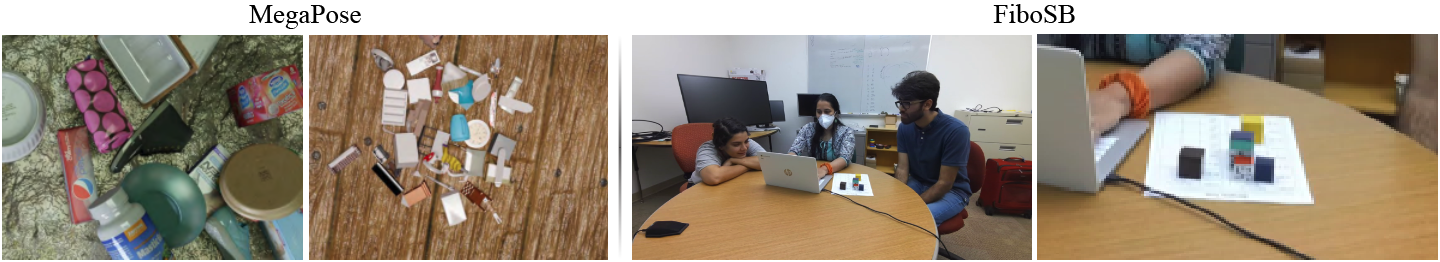}
   \caption[TrainingImgComparison]{\jung{Training image comparison between MegaPose and our FiboSB dataset. The left two images are from the synthetic training data of MegaPose \cite{labbe2022megapose}, while the images on the right represent our FiboSB training data. The right-most image under FiboSB is zoomed in for illustration, and evaluations were conducted on the original images.}}
   \label{fig:TrainingImgComparison}
\end{figure}

Recently, extensive research has shown the feasibility of an AI agent for collaborative groups in K-12 education \jff{\cite{houde2025controlling,rizvi2023investigating}}. However, these breakthroughs are typically driven by dialogic understanding \jff{\cite{grenander2021deep,park2024empowering}} and largely ignore physical interactions across students and physical objects in these real-world settings. This gap is especially visible in collaborative activities that require hands-on manipulation, where solutions often rely on limited modalities such as verbal and textual input data rather than spatial and temporal clues. As a result, any AI system can only receive partial insight into group dynamics and task progress, limiting its ability to provide timely interventions and personalized support. 

\sm{From an educational context}, effective collaborative learning involves not only individual problem-solving but also group cooperation and physical interaction. Interactions during teamwork can include pointing, assembling, and manipulating objects to describe and prove concepts for elaborating deeper understanding between peers. Here, an AI-powered agent can play a crucial role; by monitoring how students handle and position objects, an agent can offer context-aware guidance as feedback and assessment during group work. Particularly in K-12 settings, where students vary in their developmental levels and learning styles, an agent that ``sees'' and ``understands'' these interactions can adapt instructions, prompt collaborative discussions, and highlight suggestions by each student’s progress in the task.

Beyond facilitating collaboration, a precise awareness of physical space and object relationships is essential for understanding many concepts in education, especially in spatial reasoning. 6D pose estimation \sm{entails the ability to track} the 3D position and orientation of objects\footnote{The term 6D comes from the need to predict the 3D translation and the 3D rotation of the object's pose}, enabling detailed monitoring, such as how objects move or rotate and how they relate to each other in the environmental setting. This capability is particularly beneficial for younger or lower-grade students, who often depend on visual and tactile experiences to understand geometrical concepts: distance, shape, and scale. By recognizing real-time positions and orientations of objects, we can better measure students’ performance, observe their decision-making processes, and provide elaborate feedback.

\smm{Educational studies have started incorporating tactile objects and 3D elements in lessons, benefiting students who lack spatial reasoning skills \cj{\cite{AMIR2020e04052,unal2009differences}},} providing a compelling opportunity to integrate 6D pose estimation into collaborative learning. In this paper, we investigate the potential of 6D pose estimation in supporting K-12 collaborative tasks 
and outline how this technique can enhance spatial understanding, foster teamwork, and ultimately improve the overall learning experience.

To complement and extend the modalities studied in prior work \cite{bradford2023automatic,khebour-etal-2024-common,venkatesha2024propositional}, we introduce a novel 6D pose dataset called Fibonacci Small Blocks (FiboSB), which is adapted from the Weight Task Dataset (WTD) \cite{khebour2024text}. 


In summary, the key contributions of this study are as follows.
\begin{itemize}
\item Collection of a novel 6D pose dataset specifically designed for educational settings involving collaborative group tasks.
\item Exploration of baseline performance of multiple state-of-the-art 6D pose methods on our dataset, highlighting the distinct challenges posed by collaborative group scenarios. 
\end{itemize}

\section{Dataset: Fibonacci Small Blocks (FiboSB)}
\jf{To demonstrate the potential of using 6D pose estimation for collaborative group work in an educational setting, we introduce the FiboSB dataset. FiboSB is based on the Weights Task Dataset (WTD) \cite{khebour2024text}, which involves a group of triads \sm{interacting with} six colored blocks \cj{  (two $1.5$ $inch^{3}$ and four $2.0$ $inch^{3}$)} and a weight scale to determine the weights of each block. The group is initially given the weight of one block, and then they are tasked with finding the weights of the rest of the colored blocks, which follows a Fibonacci sequence. In FiboSB, we annotate the 6D poses of the colored blocks in the WTD so that we can train and evaluate 6D pose estimation models.}

\subsubsection{Predicaments in FiboSB}


 Detecting small blocks in the collaborative group work scene is a challenge (the two images on the right of Figure \ref{fig:TrainingImgComparison}). Occlusions among the blocks make the visibility of each block worse, as shown in the last column of Figure \ref{fig:TrainingImgComparison}, which frequently occur during the group task. \cjj{The FiboSB dataset contains 25,381 annotated frames across 10 groups, with the number of frames per group ranging from 1,257 to 3,967. From the annotated frames, 133,263 object instances were annotated in total. On average, each frame contains 5.25 objects, indicating that multiple colored blocks frequently appear together. \cjj{The blocks are often placed close to other blocks, leading to frequent occlusions that increase the complexity of the dataset.} This obstacle leads to difficulties in estimating the exact object positions and their orientations. In addition, one pixel off on an annotation or prediction results in a huge error for small objects. For these reasons, recognizing the blocks and estimating their precise 6D pose predictions are critical requirements in the FiboSB dataset. }

\subsubsection{\jf{FiboSB vs. Other 6D Pose Datasets}} \label{sec:fibosb_vs_others}
To our knowledge, FiboSB is the first 6D pose dataset aimed at collaborative tasks in educational settings.
Many 6D pose datasets \cj{\cite{brachmann2014learning,calli2015ycb,guo2023handal}} are targeted at various obstacles such as textureless and transparent objects \cj{\cite{hodan2017t,jung2024housecat6d}}. 
Figure \ref{fig:TrainingImgComparison} shows examples of synthetic training data for MegaPose; the images contain everyday objects, such as clocks, bottles, toys, etc., which have substantially different colors and shapes. This is in stark contrast to our dataset, where the blocks are objects used in a collaborative group task and are not the focus of the WTD. 

\section{Methodology} \label{sec:methodology}
In this study, we explore state-of-the-art methods as baselines to evaluate the FiboSB dataset. \sm{Most} 6D pose estimation methods use a two-stage pipeline: (1) object detection, which provides 2D bounding boxes for predicted objects, and (2) 6D pose estimation, which predicts the objects' 3D translations and orientations based on the interest regions from the first stage. 

\textbf{Object Detection metrics:} We evaluated the initial stage of the state-of-the-art 6D pose estimation approaches, which is the object detection modules, by employing the $mAP_{50}$ metric (mean of Average Precision (AP) at Intersection of Union (IoU) threshold of 0.5) \cjj{\cite{everingham2015pascal,everingham2010pascal}. The values of $mAP_{50}$ range between 0\% to 100\%, and a higher value represents better performance.}

\textbf{6D Pose Estimation metrics:} 
Since an object's position in 3D space is decided by its translation and rotation, the 6D pose estimation technique provides spatial information of the object. The object detection module delivers 2D spatial information for each object. Next, the 6D pose estimation module predicts corresponding translation and rotation in the 3D coordinate system.

\cjj{For evaluation metrics of 6D pose estimation, we employed $Proj2D$ \cite{he2022onepose++} and $ADD-S$ \cite{zhang20246d}. While $Proj2D$ assesses differences in 2D space, the $ADD-S$ metric quantifies errors in 3D space.}

\subsubsection{\sm{Experimental Details}}

To establish baseline performance on FiboSB, \sm{we train SOTA RGB-based 6D pose methods namely, CosyPose \cj{\cite{labbe2020cosypose}}, RADet \cj{\cite{li2020radet}}, \cj{and YOLOX-m-6D \cite{maji2024yolo}}. Furthermore, we evaluate MegaPose \cite{labbe2022megapose} to determine whether zero-shot-based approaches can provide reliable estimations for previously unseen objects in our collaborative setting. CosyPose, RADet, and YOLOX-m-6D are trained from scratch on our dataset using a group-wise split; groups 9 and 10 were assigned to the test set, and the remaining groups to the training set. The predictions are then assessed using the appropriate metrics as stated above}. 


\section{Results}

Although the SOTA methods show robustness on other (traditional) 6D pose datasets, the baselines were unreliable on our collaborative setting for both object detection and 6D pose estimation modules. \cjj{We trace the reason for failure and then address the object detection issues with DETR and YOLO11-x \cite{carion2020end,yolo11_ultralytics}}. 


\subsubsection{\jf{Initial Evaluation of 6D Pose Estimation}} \label{sec:results_6D_pose_estimation}

\cjj{We discovered that all of the models, except for MegaPose, failed to make any predictions during evaluation. Megapose received \cj{a poor score on the \emph{ADD-S} metric ($0.16$) with a threshold of 0.1 diameter}, which indicates that it struggled to make fine-grained predictions of the colored blocks using ground-truth bounding box information (Table \ref{tab:all_metrics_megapose}); MegaPose has the most trouble at estimating pose for the yellow blocks with an average error in 3D distance of $157.53 mm$ from the ground truth labels.} 

\subsubsection{\jf{Why is 6D Pose Failing?}} \label{sec:why_6dpose_failed}

After obtaining the results from the evaluation of the 6D pose models, we were left with the burning question: what is causing CosyPose, RADet, and YOLOX-m-6D to not make any predictions? We decided to take a step back and analyze the multi-stage architectures of these models from the beginning, starting with the initial stage that performs object detection.

\cjj{We first evaluated if there were any issues with our 6D pose implementations. Labbé et al. demonstrated that their method, MegaPose, achieves robust pose estimation performance, with average recall scores of 90.5 and 88.9 for ADD (0.1d) and Proj2D (5px) respectively, outperforming the Multi-Path method on the ModelNet dataset (RGB) \cite{labbe2022megapose}. It provides a good contrast to the results on FiboSB, suggesting that the issue was related to the increased difficulty of the FiboSB dataset.}

\begin{table}[b]
\centering
\caption{Overall metrics for 6D pose estimation module of MegaPose.}
\label{tab:all_metrics_megapose}
\renewcommand{\arraystretch}{1}
\setlength{\tabcolsep}{3pt}
\begin{tabular}{lccccccc}
\toprule
\textbf{Metric} & \textbf{Red} & \textbf{Yellow} & \textbf{Green} & \textbf{Blue} & \textbf{Purple} & \textbf{Brown} & \textbf{Overall} \\
\midrule
3D Distance (mm) $\downarrow$ & 105.98 & 157.53 & 106.86 & 87.00 & 86.24 & 73.89 & \textbf{106.17} \\
Proj2D (px) $\downarrow$       & 18.88   & 23.27    & 25.58   & 18.83  & 24.37  & 21.52  & \textbf{22.11}  \\
ADD-S (0.1d) $\uparrow$        & 0.17    & 0.12     & 0.08    & 0.17   & 0.15   & 0.42   & \textbf{0.16}   \\
\bottomrule
\end{tabular}
\end{table}

Next, we performed an ablation of the 6D pose methods. We realized that the object detection modules were failing, i.e., they did not detect any of the blocks in the test set and thus caused the 6D pose stage of the models to make no predictions whatsoever. \cjj{We evaluated the object detection modules individually and found that they received astonishingly low $mAP_{50}$ scores--CosyPose with 0.004, RADet with 0.000, and YOLOX-m-6D with 0.005.}
\subsubsection{\jung{\jf{Addressing and Analyzing Object Detection Problem}}} \label{sec:evaluating_object_detection_yolo11}

\jung{After pinpointing the failure of the 6D pose estimation models as originating from the object detection modules, we explored more sophisticated object detection models such as DETR and YOLO11-x \cite{carion2020end,yolo11_ultralytics} to verify our FiboSB dataset. The DETR and YOLO11-x methods achieved 0.706 and 0.898 on the $mAP_{50}$ metric respectively. Notably, the DETR model trained from scratch also failed on our dataset like the baselines. Table \ref{tab:detection} shows the results of YOLO11-x using the $mAP_{50}$ metric. We assumed that YOLO11-x performed better because of additional data augmentations and multi-scale techniques. This experiment further proved to us that a more sophisticated object detection model is capable of accurately detecting the colored blocks.}

\begin{table}[]
\centering
\cjj{\caption{The performances of fine-tuned YOLO11-x with the $mAP_{50}$ metric under our additional data. We assigned a larger portion of groups to the validation and test sets compared to the initial experiments (Section \ref{sec:methodology}). Specifically, groups 1 to 4, groups 5 to 7, and groups 8 to 10 were distributed to train, validation, and test sets respectively. The first three rows compare the performance across groups in the test set. The last rows represent the overall results on the test set.}}
\label{tab:detection}
\renewcommand{\arraystretch}{1.2}
\setlength{\tabcolsep}{3pt}
\begin{tabular}{lccccccc}
\toprule
 \textbf{Additional Data} & \textbf{Red} & \textbf{Yellow} & \textbf{Green} & \textbf{Blue} & \textbf{Purple} & \textbf{Brown} & \textbf{All} \\
\midrule
Group 8 & 0.995 & 0.995 & 0.995 & 0.841 & 0.995 & 0.991 & 0.969 \\
Group 9 & 0.995 & 0.995 & 0.995 & 0.782 & 0.000 & 0.456 & 0.704 \\
Group 10 & 0.887 & 0.992 & 0.905 & 0.893 & 0.896 & 0.993 & 0.928 \\
\midrule
FiboSB & 0.961 & \textbf{0.990} & 0.967 & 0.851 & 0.765 & 0.852 & 0.898 \\
\bottomrule
\end{tabular}
\end{table}

\section{Discussion and Conclusion}
\jung{6D pose estimation provides essential spatial context between students and objects in collaborative settings for AI agents. The agents enable us to trace student performance and infer the reasons behind object movements for immediate and objective feedback on collaborative group tasks such as in  \cite{vanderhoeven2025trace}. These capacities foster teamwork and improve learning experiences, and instructors can focus on higher-level advising and personalized support while the agent handles assessments and tracking the group process. 
} As an initial study exploring the advantages, we focused on a simplified problem, the Fibonacci weight task, to explore the performances of the current state-of-the-art methods on our dataset. 
\sm{\cjj{Based on current SOTA 6D pose estimation models, our findings reveal that existing object detection modules within these models lack the capabilities to even detect small objects in our collaborative setting.} \cj{Moreover, the 6D pose estimation module still contains gaps to accomplish the fine-grained predictions.} New 6D pose foundation models need to be developed that are able to detect small objects \cj{and produce precise 6D pose estimations} in educational settings, collaborative or otherwise, such that an AI agent uses object detections to understand the problem at hand and eventually provide optimal guidance.}

\section{Acknowledgment}
\jung{This material is based in part upon work supported by the National Science Foundation (NSF) under
subcontracts to Colorado State University on award DRL 2019805 (Institute
for Student-AI Teaming), and by Other Transaction award HR00112490377 from the U.S. Defense Advanced Research Projects Agency (DARPA) Friction for Accountability in Conversational Transactions
(FACT) program. Approved for public release, distribution unlimited. Views expressed herein do not
reflect the policy or position of the National Science Foundation, the Department of Defense, or the U.S.
Government. All errors are the responsibility of the authors.}

%
%
%
\bibliographystyle{splncs04}
\bibliography{allbibs}

\end{document}